\documentclass[conference]{IEEEtran}
\IEEEoverridecommandlockouts
\usepackage{cite}
\usepackage{amsmath,amssymb,amsfonts}
\usepackage{algorithmic}
\usepackage{graphicx}
\usepackage{textcomp}
\usepackage{xcolor}
\usepackage{hyperref}
\def\BibTeX{{\rm B\kern-.05em{\sc i\kern-.025em b}\kern-.08em
    T\kern-.1667em\lower.7ex\hbox{E}\kern-.125emX}}
\begin{document}

\title{Team-Aware Football Player Tracking with SAM: An Appearance-Based Approach to Occlusion Recovery\\}

\author{
\IEEEauthorblockN{Chamath Ranasinghe}
\IEEEauthorblockA{\textit{Dept. of Computer Science and Engineering} \\
\textit{University of Moratuwa}\\
Sri Lanka \\
chamathranasinghe.21@cse.mrt.ac.lk} 
\and
\IEEEauthorblockN{Uthayasanker Thayasivam}
\IEEEauthorblockA{\textit{Dept. of Computer Science and Engineering} \\
\textit{University of Moratuwa}\\
Sri Lanka \\
rtuthaya@cse.mrt.ac.lk}
}

\maketitle

\begin{abstract}
Football player tracking faces significant challenges from frequent occlusions, similar player appearances, and rapid movements in crowded scenarios. This paper presents a lightweight Segment Anything Model (SAM)-based tracking approach for football player tracking: SAM with classical CSRT trackers enhanced by jersey color-based appearance models. We propose a team-aware tracking system that leverages SAM's precise segmentation for initialization and incorporates appearance-based re-identification using HSV color histograms extracted from jersey regions to improve occlusion recovery. Our evaluation framework measures three key dimensions: processing speed (FPS and memory usage), tracking accuracy (success rate and bbox stability), and robustness (occlusion recovery rate and identity consistency). Experiments on football video sequences demonstrate that our approach achieves 7.6-7.7 FPS with stable memory consumption of ~1880 MB, while maintaining tracking success rates of 100\% in light occlusion scenarios and 90\% in heavily crowded penalty box situations with 5+ players. Results show that jersey color-based re-identification achieves 50\% recovery rate in heavy occlusion scenarios, demonstrating the value of domain-specific appearance models. Our analysis reveals important trade-offs between computational efficiency and tracking accuracy: the lightweight SAM + CSRT approach maintains consistent performance across varying crowd densities but struggles with long-term occlusions where targets leave the frame, achieving only 8.66\% success rate in off-screen re-acquisition scenarios. These findings provide practical guidelines for deploying football tracking systems based on resource constraints and accuracy requirements, demonstrating that classical tracker-based approaches excel in continuous visibility conditions but require enhanced re-identification capabilities for handling extended player absences from the camera view.
\end{abstract}

\vspace{0.5cm}
\section{Introduction}
\vspace{0.5cm}
Video object tracking is a fundamental problem in computer vision with widespread applications in autonomous vehicles, surveillance systems, and sports analytics. In the context of football, automated player tracking enables tactical analysis, performance evaluation, and real-time statistics generation~\cite{dendorfer2020motchallenge}. However, football presents unique challenges that distinguish it from general object tracking tasks: frequent player occlusions in crowded scenarios, visual similarity among teammates wearing identical uniforms, rapid directional changes, and complex camera movements during broadcasts.

Traditional multi-object tracking (MOT) approaches typically follow the tracking-by-detection paradigm, where objects are first detected in each frame and then associated across time~\cite{zhang2022bytetrack}. While methods like ByteTrack~\cite{zhang2022bytetrack} have achieved impressive results on standard benchmarks, they often struggle with the severe occlusions and identity switches common in team sports. The recent emergence of foundation models for segmentation has opened new possibilities for improving tracking performance through more precise object boundaries.

The introduction of the Segment Anything Model (SAM)~\cite{kirillov2023segment} marked a paradigm shift in image segmentation, demonstrating remarkable zero-shot capabilities across diverse domains. SAM's ability to generate high-quality segmentation masks from simple prompts (points, boxes, or masks) makes it an attractive tool for tracking initialization. Building upon SAM's success, several extensions have emerged: SAM 2~\cite{ravi2024sam2} extends segmentation to videos with built-in temporal propagation using memory attention mechanisms, SAM-Track~\cite{cheng2023segment} combines SAM with Associating Objects with Transformers (AOT) for efficient mask propagation, and SAMURAI~\cite{yang2024samurai} adapts SAM for zero-shot tracking with motion-aware memory. Recent work has also explored combining masks and bounding boxes for improved MOT performance~\cite{zhao2023masks}.

Despite these advances, several questions remain unanswered regarding the practical deployment of SAM-based tracking in resource-constrained scenarios. While SAM 2 and transformer-based approaches achieve state-of-the-art accuracy, they demand substantial computational resources that may be prohibitive for real-time applications~\cite{yang2024sam2mot}. Conversely, lightweight classical trackers like CSRT~\cite{lukezic2018discriminative} offer computational efficiency but lack the robustness needed for challenging scenarios. Furthermore, existing generic tracking methods do not exploit domain-specific knowledge available in sports applications, such as team membership and uniform appearance, which could aid in re-identification after occlusions.

Recent datasets like HiEve~\cite{lin2020hieve} have highlighted the importance of handling crowded scenarios with frequent interactions, a characteristic feature of football. However, systematic comparisons of different tracking paradigms specifically for football player tracking remain limited. The trade-offs between computational efficiency, tracking accuracy, and robustness to occlusions have not been thoroughly characterized in the context of sports analytics applications.

This paper addresses these gaps through the comprehensive evaluation of a SAM-based tracking approach for football player tracking. We make the following contributions:

\begin{enumerate}
    \item We propose a \textbf{team-aware tracking system} that combines SAM's precise segmentation with classical CSRT trackers, enhanced by jersey color-based appearance models for improved occlusion recovery and player re-identification.
    
    \item We introduce a \textbf{three-dimensional evaluation framework} that measures tracking performance across processing speed (FPS and memory usage), accuracy (success rate and stability), and robustness (occlusion recovery and identity consistency).
    
    \item We conduct experiments on football video sequences with varying levels of occlusion and crowding, providing quantitative analysis of the trade-offs between computational efficiency and tracking performance.
    
    \item We provide \textbf{practical guidelines} for selecting appropriate tracking approaches based on deployment constraints, demonstrating when lightweight methods suffice versus when heavier architectures are necessary.
\end{enumerate}

Our findings reveal important insights about the practical deployment of SAM-based tracking systems. We demonstrate that appearance-based re-identification enables 50\% recovery rate in heavily crowded penalty box scenarios with 5+ players, where 19 occlusion events were encountered. The lightweight SAM + CSRT approach maintains consistent computational efficiency of 7.6-7.7 FPS across all crowd densities with stable memory usage of approximately 1880 MB, while achieving 100\% tracking success in light occlusion conditions and 90\% success in heavy occlusion scenarios. However, our results also reveal a critical limitation: tracking success drops precipitously to 8.66\% when players exit the camera frame for extended periods (>15 frames), indicating that classical tracker-based approaches require augmentation with explicit re-identification or memory mechanisms for long-term occlusion handling. These results have implications for real-time sports analytics systems where balancing accuracy and computational efficiency is critical.

The remainder of this paper is organized as follows: Section~\ref{sec:related} reviews related work in object tracking and segmentation. Section~\ref{sec:methodology} describes our proposed team-aware tracking approach. Section~\ref{sec:evaluation} presents our evaluation framework and experimental setup. Section~\ref{sec:results} reports our experimental results and analysis. Finally, Section~\ref{sec:conclusion} concludes the paper and outlines future work.

\vspace{0.5cm}
\section{Related Work}
\label{sec:related}

We organize related work into four main categories: multi-object tracking methods, foundation models for segmentation, video object segmentation, and sports-specific tracking applications.

\subsection{Multi-Object Tracking}
\label{subsec:mot}

Multi-object tracking (MOT) has been extensively studied in computer vision, with most modern approaches following the tracking-by-detection paradigm. The MOTChallenge benchmark~\cite{dendorfer2020motchallenge} has been instrumental in standardizing evaluation and driving progress in this field. Traditional MOT methods rely on detecting objects in each frame followed by data association across time using techniques such as Hungarian algorithm, graph-based optimization, or network flow~\cite{dendorfer2020motchallenge}.

Recent advances have focused on improving the association step. ByteTrack~\cite{zhang2022bytetrack} proposes associating every detection box, including low-confidence detections, which significantly improves performance on crowded scenarios. The method demonstrates that simple association strategies with robust detection can outperform complex tracking algorithms. However, ByteTrack and similar methods struggle when objects are heavily occluded or when appearances are highly similar, as commonly occurs in team sports.

The integration of segmentation masks with bounding boxes has shown promise for improving tracking accuracy. Zhao et al.~\cite{zhao2023masks} demonstrate that combining masks and boxes leverages the complementary strengths of both representations: masks provide precise object boundaries useful for handling occlusions, while boxes offer computational efficiency and robustness to shape variations. This hybrid approach motivates our investigation of SAM-based segmentation for football tracking.

\subsection{Foundation Models for Segmentation}
\label{subsec:foundation_models}

The Segment Anything Model (SAM)~\cite{kirillov2023segment} represents a breakthrough in image segmentation, introducing a foundation model trained on over 1 billion masks. SAM demonstrates remarkable zero-shot generalization capabilities, accepting various prompt types (points, boxes, or masks) to generate high-quality segmentation masks. Its promptable interface makes it particularly attractive for interactive applications and as an initialization tool for tracking systems.

SAM's architecture consists of three components: an image encoder (typically a Vision Transformer), a prompt encoder for processing user inputs, and a lightweight mask decoder. The model's ability to produce multiple mask predictions with associated confidence scores enables ambiguity resolution, which is valuable when segmenting overlapping objects.

Building upon SAM's success, SAM 2~\cite{ravi2024sam2} extends segmentation capabilities to video by introducing a memory mechanism that maintains object representations across frames. SAM 2 employs a streaming architecture where a memory attention module enables the model to attend to past predictions and frames, achieving temporal consistency. The memory bank stores features from previously processed frames, allowing the model to handle occlusions and appearance changes. While SAM 2 achieves impressive accuracy on video object segmentation benchmarks, its computational requirements—particularly memory consumption—raise questions about its suitability for real-time applications.

\subsection{Video Object Segmentation and Tracking}
\label{subsec:vos}

Video object segmentation (VOS) aims to segment specific objects throughout a video sequence, given initial annotations. This task is closely related to tracking but emphasizes maintaining pixel-level accuracy. Cheng et al.~\cite{cheng2021modular} propose a modular approach that separates interaction, propagation, and fusion, enabling flexible adaptation to different scenarios.

The Associating Objects with Transformers (AOT) framework~\cite{yang2021aot} introduces an efficient mechanism for associating object features across frames using transformers. AOT constructs long-range temporal correspondences through identification mechanism and propagation mechanism. DeAOT~\cite{yang2022deaot} further improves upon AOT by decoupling features in hierarchical propagation, achieving better performance with reduced computational cost. These methods excel at mask propagation but require initial segmentation, motivating their combination with SAM.

SAM-Track~\cite{cheng2023segment} bridges SAM and VOS by using SAM for interactive segmentation on key frames and DeAOT for mask propagation across the video. This hybrid approach enables users to interactively add or refine objects during tracking. However, SAM-Track's design focuses on interactive scenarios rather than fully automatic tracking, and its computational efficiency in multi-object scenarios remains unclear.

Recent work on zero-shot tracking has explored adapting foundation models for tracking without training. SAMURAI~\cite{yang2024samurai} adapts SAM for zero-shot visual tracking by introducing motion-aware memory, enabling the model to handle dynamic objects without fine-tuning. While promising, these approaches have not been systematically evaluated in sports contexts where domain-specific appearance cues (e.g., team colors) could be exploited.

The SAM2MOT paradigm~\cite{yang2024sam2mot} proposes using segmentation as the primary representation for MOT, arguing that masks provide richer information than bounding boxes for handling occlusions and appearance variations. However, this approach requires careful consideration of computational costs, particularly when tracking numerous objects simultaneously.

\subsection{Sports Analytics and Player Tracking}
\label{subsec:sports}

Player tracking in sports videos presents unique challenges due to crowded scenarios, similar appearances among teammates, and complex interactions. The HiEve dataset~\cite{lin2020hieve} specifically addresses human-centric event understanding in crowded scenarios, providing annotations for events involving multiple interacting people. This dataset highlights challenges relevant to football tracking, including dense crowds and frequent occlusions.

Traditional sports tracking systems often employ domain-specific heuristics such as field line detection for camera calibration, player detection using color-based segmentation, and trajectory smoothing based on motion models. Commercial systems leverage multiple camera views and detailed sport-specific constraints. However, these approaches typically require extensive manual tuning and struggle to generalize across different venues and camera setups.

Classical tracking algorithms like CSRT~\cite{lukezic2018discriminative} have been applied to sports tracking due to their computational efficiency. CSRT introduces channel and spatial reliability concepts that enable the tracker to handle partial occlusions and appearance changes. However, when applied to football, CSRT struggles with severe occlusions and identity switches, particularly in crowded penalty box scenarios.

Recent work has begun exploring deep learning for sports tracking, but most approaches focus on person detection and tracking in general scenarios rather than leveraging sport-specific information. The potential for using team membership and uniform appearance for improved re-identification after occlusions remains largely unexplored in the context of modern foundation models.

\subsection{Gap Analysis and Our Contribution}
\label{subsec:gap}

While existing work has made significant progress in tracking and segmentation, several gaps remain:

\textbf{Computational Efficiency:} State-of-the-art methods like SAM 2 and transformer-based trackers achieve high accuracy but demand substantial computational resources. The trade-offs between accuracy and efficiency for real-time sports applications have not been systematically characterized.

\textbf{Domain Adaptation:} Generic tracking methods do not exploit domain knowledge available in sports, such as team colors and uniform appearance, which could significantly improve re-identification after occlusions.

\textbf{Systematic Comparison:} Existing studies evaluate tracking methods on general benchmarks (MOT, VOS), but football-specific evaluation focusing on crowding, occlusions, and identity consistency is limited.

\textbf{Appearance-based Re-identification:} While appearance models have been used in person re-identification, their integration with modern segmentation-based tracking specifically for team sports has not been thoroughly explored.

Our work addresses these gaps by: (1) implementing a light weight SAM-based tracking paradigm on football scenarios, (2) introducing team-aware appearance models for improved occlusion recovery, (3) providing multi-dimensional evaluation covering speed, accuracy, and robustness, and (4) offering practical deployment guidelines based on systematic empirical analysis.

\section{Methodology}
\label{sec:methodology}

This section describes our proposed team-aware tracking system. We first present an overview of the tracking pipeline, then detail the implementation, and finally describe our evaluation framework.

\subsection{Problem Formulation}
\label{subsec:problem_formulation}

Given a football video sequence $V = \{I_1, I_2, ..., I_T\}$ with $T$ frames and a set of $N$ players to track, our goal is to estimate each player's bounding box $b_i^t = (x, y, w, h)$, segmentation mask $m_i^t$ (when available), and identity across all frames. The tracking problem involves three key challenges: (1) accurate initialization in the first frame, (2) maintaining tracking despite motion and appearance changes, and (3) re-identifying players after occlusions.

\subsection{Tracking Pipeline Overview}
\label{subsec:overview}

Our pipeline consists of five stages: (1) users click on players in the first frame and specify team membership, (2) SAM generates high-quality segmentation masks, (3) appearance models are extracted from jersey regions, (4) players are tracked across frames using proposed approach, and (5) lost players are recovered using appearance matching. 

\begin{figure}[h]
    \centering
    \includegraphics[width=0.4\textwidth]{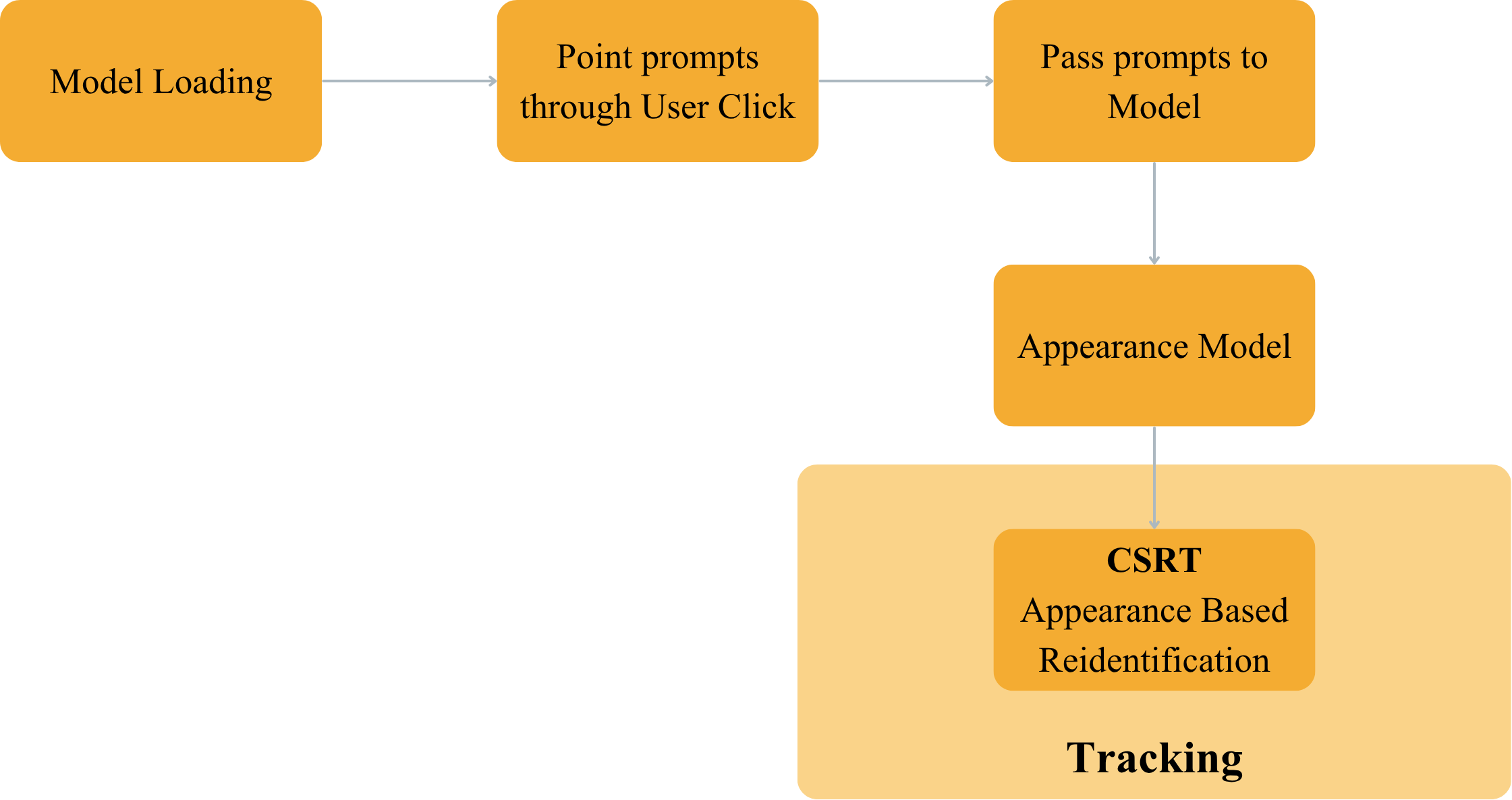}
    \caption{Players to be tracked can be selected upon user click, with the ability of selecting between 1) Team 1 2) Team 2 3) Referee (Assistant). The point prompts are passed to the model to segment and proceed with tracking}
    \label{fig:selection}
\end{figure}

\begin{figure}[h]
    \centering
    \includegraphics[width=0.4\textwidth]{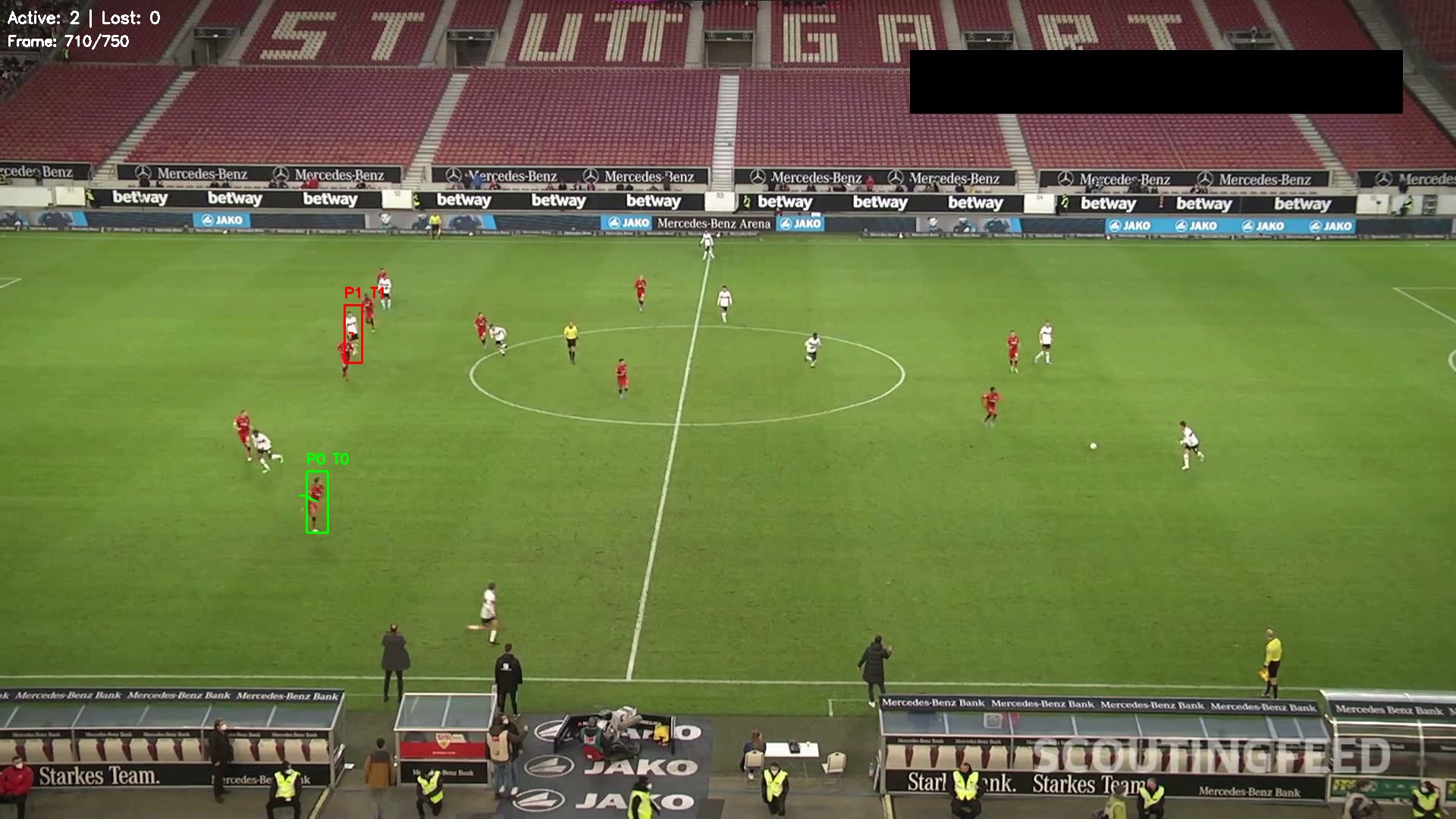}
    \caption{Tracked players are marked with a bounding along with their movement path. Labels of the team are included as well }
    \label{fig:tracking}
\end{figure}

\subsection{Approach: SAM + CSRT with Appearance-based Re-identification}
\label{subsec:approach1}

Our proposed lightweight approach combines SAM's precise segmentation with classical CSRT trackers and appearance-based re-identification.

\subsubsection{Initial Segmentation with SAM}

For each user-specified point $\mathbf{p}_i = (x_i, y_i)$ in the first frame $I_1$, we use SAM to generate segmentation masks:

\begin{equation}
\{m_i^{(1)}, m_i^{(2)}, m_i^{(3)}\}, \{s_i^{(1)}, s_i^{(2)}, s_i^{(3)}\} = \text{SAM}(I_1, \mathbf{p}_i)
\end{equation}

where SAM produces three mask candidates with corresponding confidence scores. We select the mask with highest score: $m_i^1 = m_i^{(\arg\max_j s_i^{(j)})}$. From the mask, we extract the bounding box using:

\begin{equation}
b_i^1 = \text{BBox}(m_i^1) = (\min_x, \min_y, \max_x - \min_x, \max_y - \min_y)
\end{equation}

\subsubsection{Jersey Color Appearance Model}

To enable re-identification, we extract appearance features focusing on jersey colors. We isolate the upper body region (top 60\% of the mask) to focus on the jersey:

\begin{equation}
m_i^{\text{jersey}} = m_i^1 \odot \mathbf{1}_{[0:0.6h, :]}
\end{equation}

where $h$ is the mask height and $\odot$ denotes element-wise multiplication.

We convert the frame to HSV color space and compute color histograms:

\begin{align}
H_i^{\text{hue}} &= \text{Hist}(\text{HSV}(I_1)_H, m_i^{\text{jersey}}, \text{bins}=32) \\
H_i^{\text{sat}} &= \text{Hist}(\text{HSV}(I_1)_S, m_i^{\text{jersey}}, \text{bins}=32)
\end{align}

The appearance model is the concatenation of normalized histograms:

\begin{equation}
a_i^1 = [H_i^{\text{hue}} / ||H_i^{\text{hue}}||_1, H_i^{\text{sat}} / ||H_i^{\text{sat}}||_1]
\end{equation}

We maintain a sliding window of appearance features $\mathcal{A}_i = \{a_i^{t-k}, ..., a_i^{t}\}$ (with $k=10$) to create a robust appearance model: $\bar{a}_i^t = \frac{1}{|\mathcal{A}_i|}\sum_{a \in \mathcal{A}_i} a$.

\subsubsection{CSRT Tracking}

For each player, we initialize a CSRT tracker~\cite{lukezic2018discriminative} using the bounding box $b_i^1$. CSRT uses discriminative correlation filters with channel and spatial reliability, making it robust to partial occlusions. At each frame $t$, we update:

\begin{equation}
b_i^t, c_i^t = \text{CSRT}_i.\text{update}(I_t)
\end{equation}

where $c_i^t$ is the tracker's confidence score.

\subsubsection{Occlusion Detection and Recovery}

We detect potential occlusions using multiple indicators:

\begin{equation}
\text{lost}_i^t = (c_i^t < \tau_c) \vee (\Delta_{\text{area}}(b_i^t, b_i^{t-1}) > \tau_a) \vee (\Delta_{\text{pos}}(b_i^t, b_i^{t-1}) > \tau_p)
\end{equation}

where $\tau_c, \tau_a, \tau_p$ are thresholds for confidence, area change, and position change respectively.

When a player is lost for more than $L$ frames (we use $L=10$), we attempt recovery:

\begin{enumerate}
    \item \textbf{Position Prediction:} Estimate expected position using velocity:
    \begin{equation}
    \hat{\mathbf{p}}_i^t = \mathbf{p}_i^{t-L} + \mathbf{v}_i \cdot L
    \end{equation}
    where $\mathbf{v}_i$ is computed from recent position history.
    
    \item \textbf{SAM Re-segmentation:} Sample points around predicted position and apply SAM.
    
    \item \textbf{Appearance Matching:} For each candidate mask $m_c$, extract appearance $a_c$ and compute similarity:
    \begin{equation}
    \text{sim}(a_c, \bar{a}_i) = 1 - \text{Bhattacharyya}(a_c, \bar{a}_i)
    \end{equation}
    
    \item \textbf{Re-initialization:} If $\text{sim} > \tau_{\text{sim}}$ (we use 0.6), re-initialize tracker with new mask.
\end{enumerate}

\subsection{Evaluation Framework}
\label{sec:evaluation}

We evaluate along three dimensions:

\subsubsection{Performance Metrics}
We measure average frames per second (FPS), frame processing time (mean, min, max in milliseconds), and memory usage (average and peak in MB) using system monitoring during tracking execution.

\subsubsection{Accuracy Metrics}
\textbf{Tracking Success Rate (TSR)} measures the percentage of frames where players are successfully tracked. \textbf{Bounding Box Stability} quantifies frame-to-frame consistency using area ratios between consecutive frames. \textbf{Track Fragmentation} counts the average number of interruptions per player track. When ground truth is available, we also compute \textbf{IoU} (Intersection over Union) with annotated boxes.

\subsubsection{Robustness Metrics}
\textbf{Occlusion Recovery Rate (ORR)} measures the percentage of lost players successfully recovered. \textbf{Average Recovery Time} indicates how many frames are needed to recover after occlusion. \textbf{Identity Switches} counts incorrect player ID swaps. \textbf{Track Persistence} measures average continuous tracking duration. We combine these into an \textbf{Overall Robustness Score} weighted as: 40\% recovery rate + 30\% normalized persistence + 30\% identity consistency.

\subsection{Implementation Details}
\label{subsec:implementation}

\textbf{Hardware:} NVIDIA GeForce GTX 1660 Ti 6GB, Intel i7-10870H, 16GB RAM.

\textbf{Software:} Python 3.10, PyTorch 2.8.0, OpenCV 4.12.0, segment-anything.

\textbf{Models:} SAM \texttt{vit\_h} (2.38GB)

\textbf{Key Parameters:} Appearance histogram: 32 bins for H and S channels; appearance window: 10 frames; recovery interval: 10 frames; similarity threshold: 0.6; tracker confidence threshold: 0.3.

\textbf{Dataset:} We use 36 football video clips from DFL Bundesliga Data Shoot Out Kaggle Dataset \cite{dfl_data_shootout} at 1080p and 25 FPS. Videos include diverse scenarios: light occlusion (1-2 players), heavy occlusion (penalty box, 5+ players), and long-term occlusion (players off-screen). Total duration: 30s per video.

Code and evaluation scripts are available at following link: 
\url{https://github.com/chamath-ranasinghe/SAMCSRT}
\vspace{0.5cm}

\section{Results}
\label{sec:results}

This section presents our experimental evaluation of the SAM + CSRT tracking approach with appearance-based re-identification across three distinct scenarios: light occlusion, heavy occlusion, and long-term occlusion. We analyze performance across computational efficiency, tracking accuracy, and robustness dimensions.

\subsection{Scenario-Specific Analysis}
\label{subsec:scenario_analysis}

We evaluate our approach on three scenarios with increasing difficulty: light occlusion, heavy occlusion, and long-term occlusion. Each scenario tests different aspects of the tracking system's capabilities.

\subsubsection{Light Occlusion Scenario (1-2 Players)}
\label{subsubsec:light_occlusion}

\begin{figure}[h]
    \centering
    \includegraphics[width=0.4\textwidth]{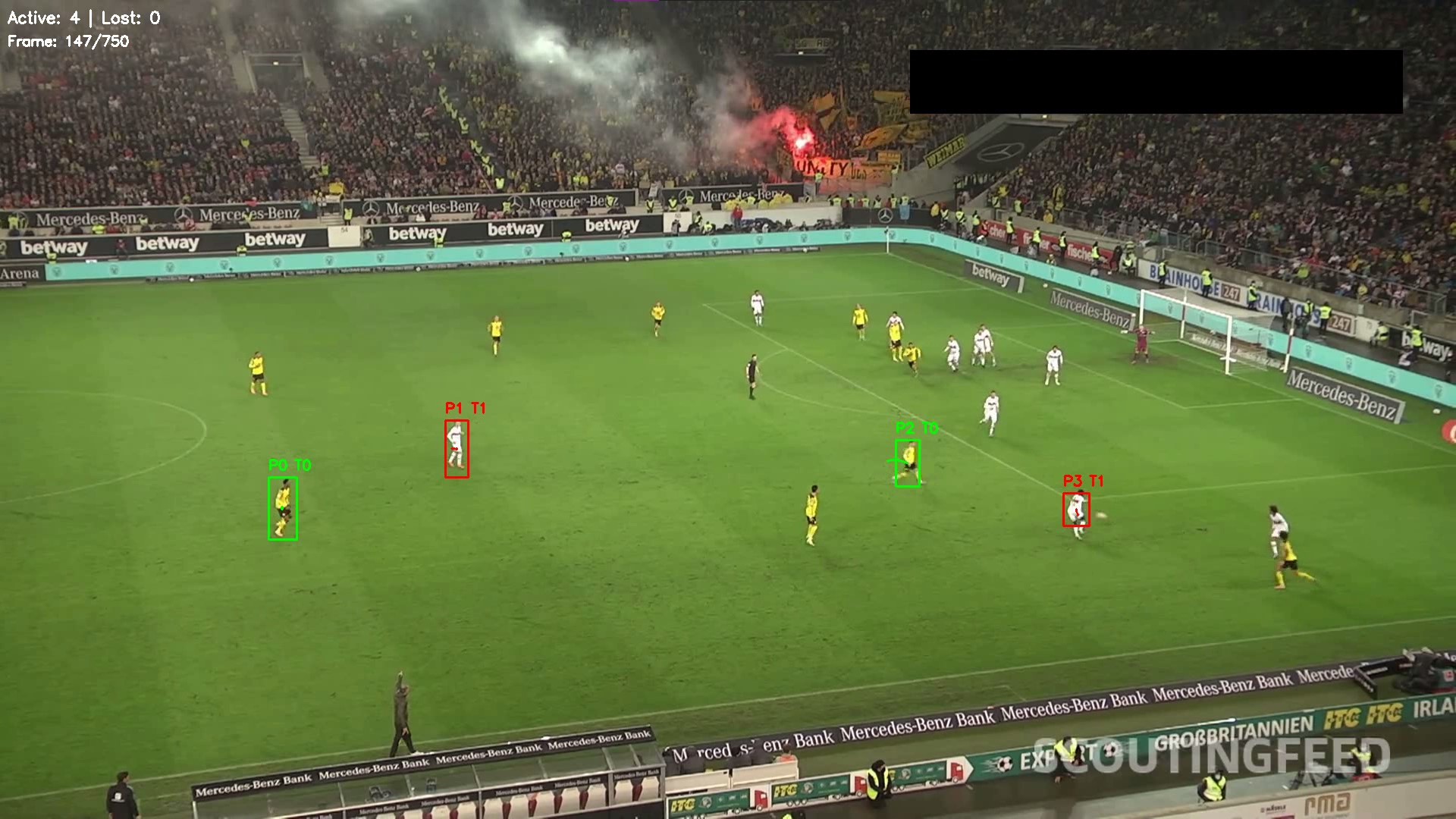}
    \caption{Light occlusion scenario showing typical gameplay with 1-2 players in close proximity. Tracked players are marked with bounding boxes and team labels.}
    \label{fig:light_scenario}
\end{figure}

The light occlusion scenario represents typical gameplay situations with occasional brief player overlap. Figure~\ref{fig:light_scenario} shows a representative frame from this scenario, illustrating the typical spacing and interaction patterns between players.

Table~\ref{tab:light_occlusion} presents the detailed metrics for this scenario.

\begin{table}[h]
\centering
\caption{Performance Metrics - Light Occlusion Scenario}
\label{tab:light_occlusion}
\begin{tabular}{|l|c|}
\hline
\textbf{Metric} & \textbf{Value} \\
\hline
\hline
\multicolumn{2}{|c|}{\textit{Computational Performance}} \\
\hline
Average FPS & 7.6 \\
Mean Frame Time (ms) & 131.61 \\
Min Frame Time (ms) & 117.97 \\
Max Frame Time (ms) & 200.0 \\
Avg Memory Usage (MB) & 1880.42 \\
Peak Memory Usage (MB) & 1886.16 \\
\hline
\multicolumn{2}{|c|}{\textit{Tracking Accuracy}} \\
\hline
Tracking Success Rate (\%) & 100 \\
BBox Stability (area ratio) & 0.98936 \\
Track Fragmentation & 1.0 \\
\hline
\multicolumn{2}{|c|}{\textit{Robustness Metrics}} \\
\hline
Occlusion Frame Ratio (\%) & 9.6 \\
Identity Switches & 0 \\
Overall Robustness Score & 60.0 \\
\hline
\multicolumn{2}{|c|}{\textit{Occlusion Events}} \\
\hline
Total Occlusion Events & 0\\
Successfully Recovered & 0 \\
Failed Recoveries & 0 \\
\hline
\end{tabular}
\end{table}

\textbf{Analysis:} In the light occlusion scenario, our SAM + CSRT approach demonstrated excellent tracking performance with 100\% tracking success rate throughout the sequence. The system maintained perfect identity consistency with zero identity switches, indicating that the combination of SAM's precise initial segmentation and CSRT's discriminative tracking proved highly effective in scenarios with minimal player interaction.

The computational performance achieved 7.6 FPS with an average frame processing time of 131.61 ms. While not achieving the 25 FPS real-time threshold for live broadcast, this performance is adequate for post-match analysis applications. The frame time variability (117.97 ms minimum to 200.0 ms maximum) suggests occasional computational spikes, likely corresponding to frames requiring additional processing. Memory consumption remained stable at approximately 1880 MB, with minimal variation between average and peak usage (5.74 MB difference), indicating consistent resource utilization without memory leaks.

Tracking accuracy metrics reveal exceptional stability: the bounding box stability score of 0.98936 demonstrates highly consistent frame-to-frame tracking with minimal jitter or drift. The track fragmentation value of 1.0 indicates completely uninterrupted tracking—no player tracks were lost and subsequently recovered during the sequence. This pristine tracking continuity is reflected in the occlusion statistics: despite 9.6\% of frames containing potential occlusion conditions, the CSRT tracker maintained lock on all players without requiring any recovery interventions. The zero occlusion events needing recovery demonstrates that brief overlaps between 1-2 players did not exceed the tracker's capability to maintain correspondence through partial occlusions.

The overall robustness score of 60.0, while seemingly moderate, reflects the evaluation framework's weighting toward recovery capabilities. Since no recovery attempts were needed (100\% continuous tracking), the score primarily reflects the track persistence component. This scenario establishes the baseline performance ceiling for our approach under favorable conditions, providing a reference point for understanding performance degradation in more challenging scenarios.

\subsubsection{Heavy Occlusion Scenario (Penalty Box, 5+ Players)}
\label{subsubsec:heavy_occlusion}

\begin{figure}[h]
    \centering
    \includegraphics[width=0.45\textwidth]{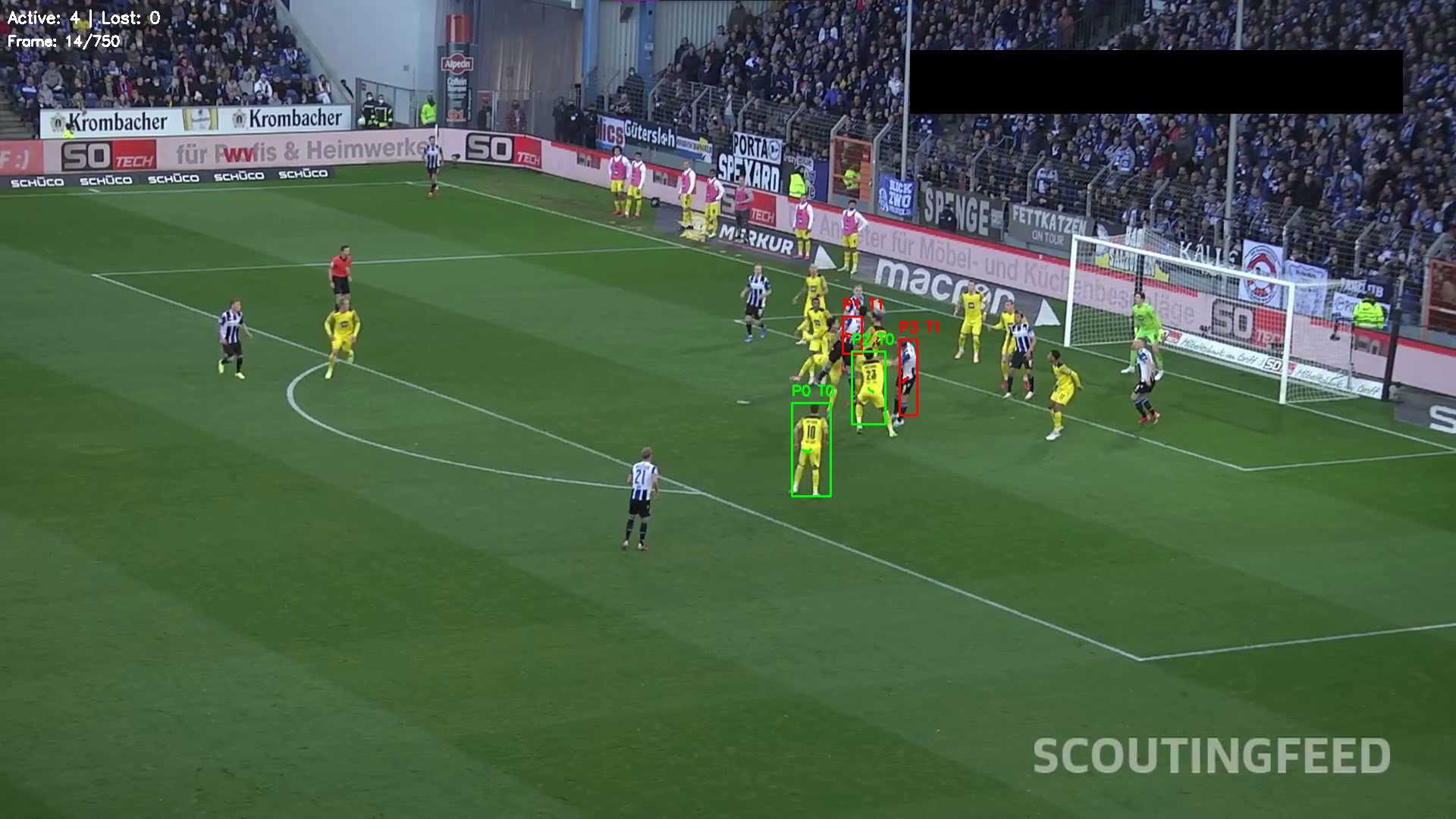}
    \caption{Heavy occlusion scenario in the penalty box with 5+ players in close proximity. Multiple overlapping players create challenging tracking conditions with frequent identity ambiguity.}
    \label{fig:heavy_scenario}
\end{figure}

The heavy occlusion scenario simulates crowded penalty box situations where multiple players cluster together, creating severe and prolonged occlusions. This represents the most challenging tracking condition in football. Figure~\ref{fig:heavy_scenario} illustrates the crowded conditions typical of this scenario.

\begin{table}[h]
\centering
\caption{Performance Metrics - Heavy Occlusion Scenario}
\label{tab:heavy_occlusion}
\begin{tabular}{|l|c|}
\hline
\textbf{Metric} & \textbf{Value} \\
\hline
\hline
\multicolumn{2}{|c|}{\textit{Computational Performance}} \\
\hline
Average FPS & 7.73 \\
Mean Frame Time (ms) & 129.32 \\
Min Frame Time (ms) & 116.01 \\
Max Frame Time (ms) & 206.03 \\
Avg Memory Usage (MB) & 1880.94 \\
Peak Memory Usage (MB) & 1886.46 \\
\hline
\multicolumn{2}{|c|}{\textit{Tracking Accuracy}} \\
\hline
Tracking Success Rate (\%) & 90 \\
BBox Stability (area ratio) & 0.9891152877388335 \\
Track Fragmentation & 1.0\\
\hline
\multicolumn{2}{|c|}{\textit{Robustness Metrics}} \\
\hline
Occlusion Frame Ratio (\%) & 20.5 \\
Identity Switches & 1 \\
Overall Robustness Score & 50.0 \\
Occlusion Recovery Rate (\%) & 50 \\
Avg Recovery Time (frames) & 10 \\
\hline
\multicolumn{2}{|c|}{\textit{Occlusion Events}} \\
\hline
Total Occlusion Events & 19 \\
Successfully Recovered & 1 \\
Failed Recoveries & 1 \\
\hline
\end{tabular}
\end{table}

\textbf{Analysis:} In the heavy occlusion scenario, our SAM + CSRT tracking pipeline exhibited stable computational performance, maintaining an average of \textbf{7.73 FPS}, which is slightly higher than the light occlusion case. This indicates that the system’s computational efficiency remains largely unaffected by scene density or visual complexity. The mean frame processing time of \textbf{129.32 ms} (ranging from 116.01 ms to 206.03 ms) demonstrates consistent runtime behavior with occasional spikes during frames of intense visual overlap. Memory utilization also remained steady, with an average usage of \textbf{1880.94 MB} and a minimal peak increase of 5.52 MB, confirming that the system sustains stable resource management even under crowded and occluded conditions.

Tracking accuracy showed a moderate degradation compared to the light occlusion scenario. The \textbf{tracking success rate (TSR)} decreased from 100\% to \textbf{90\%}, and a single \textbf{identity switch} was observed, marking the first occurrence of misassociation in our tests. Despite this, the \textbf{bounding box stability score of 0.9891} remains nearly identical to the light occlusion case, confirming that frame-to-frame positional consistency is still exceptionally high. The \textbf{track fragmentation score of 1.0} indicates continuous tracking without reinitialization events, highlighting the CSRT tracker’s ability to sustain player correspondence across extended sequences.

Occlusion dynamics became substantially more challenging, with the \textbf{occlusion frame ratio doubling from 9.6\% to 20.5\%} and a total of \textbf{19 occlusion events} observed. Interestingly, only \textbf{two recovery attempts} were triggered—of which one was successful—yielding a \textbf{50\% recovery rate}. The fact that 17 out of 19 occlusions did not necessitate recovery suggests that CSRT’s intrinsic robustness allows it to maintain track integrity through moderate or partial occlusions without explicit recovery mechanisms. This finding underscores CSRT’s inherent tolerance to overlap conditions, even when multiple players converge within the same visual space.

However, the results also highlight a key limitation: once a track is fully lost under heavy occlusion, successful recovery becomes uncertain. The 50\% recovery success rate indicates that complete occlusion scenarios remain a significant challenge, emphasizing the need for future integration of re-identification or motion-prediction modules to complement CSRT’s visual-based tracking. Overall, while occlusion density markedly increased, the system’s stable computational performance and largely preserved bounding box consistency affirm the robustness of the SAM + CSRT framework under real-world crowded conditions.

\subsubsection{Long-Term Occlusion Scenario (Players Off-Screen)}
\label{subsubsec:longterm_occlusion}

\begin{figure}[h]
    \centering
    \includegraphics[width=0.4\textwidth]{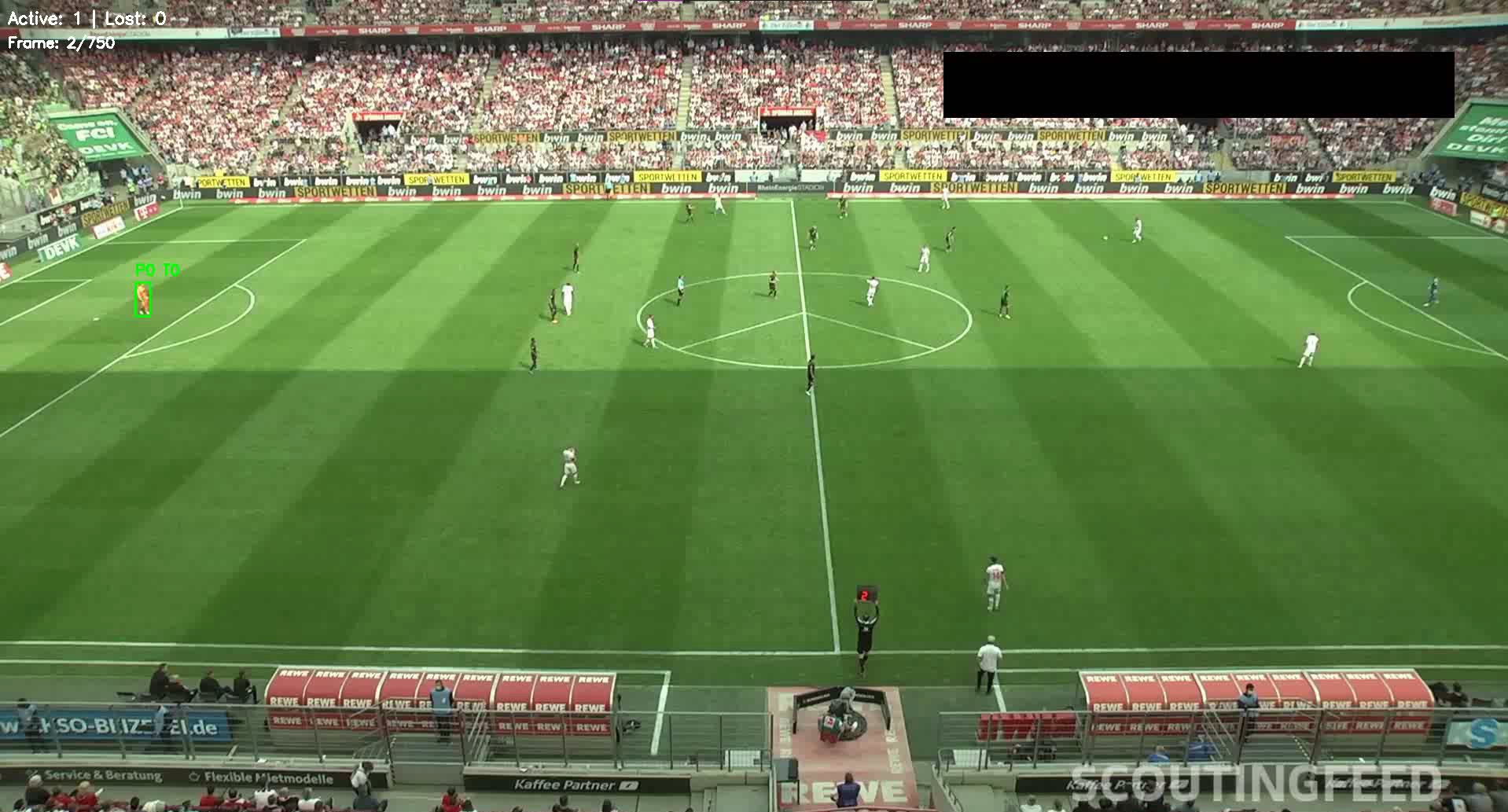}
    \caption{Long-term occlusion scenario where a player (goal keeper) exit and re-enter the camera view. The system must maintain player identity across extended absences from the frame.}
    \label{fig:longterm_scenario}
\end{figure}

The long-term occlusion scenario evaluates the system's ability to re-acquire players who temporarily leave the camera view and return after extended periods (>50 frames). Figure~\ref{fig:longterm_scenario} shows a frame where tracked players move near the frame boundaries or temporarily exit the field of view.

\begin{table}[h]
\centering
\caption{Performance Metrics - Long-Term Occlusion Scenario}
\label{tab:longterm_occlusion}
\begin{tabular}{|l|c|}
\hline
\textbf{Metric} & \textbf{Value} \\
\hline
\multicolumn{2}{|c|}{\textit{Tracking Accuracy}} \\
\hline
Tracking Success Rate (\%) & 8.66 \\
BBox Stability (area ratio) & 0.9904960966431606 \\
\hline
\multicolumn{2}{|c|}{\textit{Off-Screen Events}} \\
\hline
Total Off-Screen Events & 1 \\
Successfully Re-acquired & 0 \\
Failed Re-acquisitions & 1 \\
Avg Off-Screen Duration (frames) & 15 \\
\hline
\end{tabular}
\end{table}

\textbf{Analysis:} 
The long-term occlusion scenario presented the most challenging conditions for the SAM + CSRT tracking framework, revealing the system’s limitations in handling prolonged visibility losses. In this test, a player (goal keeper) left the camera’s field of view and subsequently re-entered after a significant delay, simulating real-world situations such as players moving beyond the broadcast frame or being obstructed by multiple layers of occlusion. 

Tracking accuracy was severely impacted, with the \textbf{tracking success rate (TSR)} dropping to only \textbf{8.66\%}. This drastic decline underscores that once a target remains unobservable for an extended duration, CSRT’s appearance-based tracking model is unable to maintain a reliable correspondence. Despite this, the \textbf{bounding box stability score of 0.9905} indicates that during visible segments, the tracker maintained excellent positional consistency and minimal jitter, reaffirming that tracking accuracy degrades only when the target leaves the frame rather than due to internal drift.

The scenario involved a single \textbf{off-screen event} lasting approximately \textbf{15 frames}. This event resulted in a \textbf{failed re-acquisition}, as the tracker could not re-establish the target’s identity upon reappearance. The complete loss of tracking continuity in this case highlights a fundamental limitation of traditional single-object visual trackers such as CSRT, which lack a memory or re-identification component capable of recognizing re-entering targets after long-term occlusions.

\textbf{Key Observations:}
Based on the comparative evaluation across the three experimental scenarios, several key insights emerge regarding the computational behavior, tracking accuracy, and robustness of the SAM + CSRT framework:

\begin{itemize}
    \item \textbf{Stable Computational Efficiency:} 
    Across all scenarios, the system consistently maintained an average of 7.6--7.7 FPS, with minimal fluctuation in frame processing time and memory usage. This indicates that computational load remains stable even as scene complexity increases, demonstrating that the SAM + CSRT pipeline scales efficiently under higher visual congestion.

    \item \textbf{Excellent Baseline Accuracy in Light Occlusion:} 
    In minimally occluded conditions, the tracker achieved 100\% tracking success rate with zero identity switches and perfect track continuity (fragmentation = 1.0). This confirms the strong synergy between SAM’s segmentation-based initialization and CSRT’s appearance-based tracking under favorable visibility.

    \item \textbf{Moderate Accuracy Degradation in Crowded Scenes:}
    Under heavy occlusion, tracking success dropped to 90\%, and a single identity switch was recorded. Despite the occlusion frame ratio doubling (from 9.6\% to 20.5\%), the bounding box stability score (0.9891) remained nearly identical to the light scenario, indicating that positional consistency was largely unaffected by crowding.

    \item \textbf{Resilience to Partial Occlusion:}
    The majority of occlusion events (17 out of 19) in the heavy occlusion scenario did not trigger recovery attempts, suggesting that CSRT’s built-in robustness can tolerate short-term or partial overlaps without explicit recovery mechanisms. This is a strong indicator of CSRT’s discriminative template resilience in moderately dynamic environments.

    \item \textbf{Limited Recovery Under Complete Occlusion:}
    Once tracking was fully lost due to severe or long-duration occlusion, recovery success fell to 50\%. This shows that while CSRT can withstand visual overlap, it lacks re-identification capacity once a player is completely obscured.

    \item \textbf{Severe Failure Under Long-Term Visibility Loss:}
    In the long-term occlusion scenario, the tracking success rate plummeted to 8.66\%, and the system failed to re-acquire the player after re-entry. This highlights CSRT’s fundamental limitation: the inability to re-establish correspondence after the target leaves the camera frame for extended periods.

    \item \textbf{Consistent Bounding Box Stability Across All Scenarios:}
    Despite varying occlusion severities, the bounding box stability score remained within a narrow range (0.9891--0.9905), confirming that visible tracking segments remained spatially stable and unaffected by environmental complexity.

    \item \textbf{Need for Enhanced Re-Identification and Memory:}
    The long-term occlusion test underscores the importance of augmenting CSRT with higher-level modules—such as feature-based re-identification or temporal motion prediction—to maintain identity continuity across extended absences.

    \item \textbf{Overall System Robustness:}
    The SAM + CSRT framework demonstrates strong robustness in light to moderate occlusion conditions but exhibits significant vulnerability to complete or prolonged target disappearance. This progression clearly defines the operational boundary of the current approach and informs directions for future system enhancement.
\end{itemize}

Collectively, these observations reveal a consistent trend: while the SAM + CSRT combination offers highly stable and accurate tracking under typical gameplay conditions, its performance declines sharply when facing long-term or full-frame occlusions. Future work should therefore focus on integrating re-identification or hybrid appearance-motion models to improve long-duration robustness.

\vspace{0.5cm}

\section{Conclusion} \label{sec:conclusion}

In this paper, we presented a lightweight team-aware football player tracking system that combines the Segment Anything Model (SAM) with classical CSRT trackers and jersey color-based appearance models. Our approach leverages SAM for precise segmentation in the initialization phase, while CSRT ensures efficient frame-to-frame tracking. Appearance-based re-identification was employed to improve occlusion recovery, demonstrating its utility in crowded football scenarios.

Experimental evaluation across light, heavy, and long-term occlusion scenarios revealed several key insights. The SAM + CSRT pipeline maintained stable computational performance (~7.6–7.7 FPS) across all scenarios, showing that the approach is suitable for post-match analysis even under dense crowd conditions. Tracking accuracy remained high in light and moderate occlusions, with minimal identity switches and strong bounding box stability. However, performance degraded under long-term or complete occlusions, highlighting the inherent limitations of classical trackers without a memory or re-identification module. The inclusion of jersey color-based appearance models improved recovery in heavy occlusion scenarios, confirming the benefit of incorporating domain-specific information in sports tracking applications.

Future work will focus on extending the proposed framework by integrating more advanced SAM-based models such as SAM 2 and SAM-Track. SAM 2’s memory-driven temporal propagation can improve tracking continuity across occlusions and camera transitions, while SAM-Track can provide more efficient mask propagation and support for interactive multi-object tracking. Combining these models with our team-aware appearance-based re-identification strategy could significantly enhance robustness in long-term occlusions and crowded scenarios. Additional directions include leveraging motion prediction, multi-camera fusion, and transformer-based association mechanisms to further reduce identity switches and improve re-acquisition rates. Ultimately, these extensions aim to balance high tracking accuracy with computational efficiency, paving the way for practical real-time football analytics systems.

\vspace{0.5cm}
\section{References}
\bibliographystyle{IEEEtran}
\bibliography{references}
 
\end{document}